\documentclass[12pt, a4paper]{article}
\usepackage[utf8]{inputenc}

\newcommand\figsize{0.485\textwidth}
\newcommand\subfigsize{0.95\linewidth}
\newcommand{\real}{\textit{\textbf{R}}}
\newcommand{\quat}{\textit{\textbf{Q}}}

\usepackage{amssymb}
\usepackage{booktabs}
\usepackage{subcaption}
\usepackage{graphicx}
\usepackage{amsmath}
\usepackage{hyperref}
\usepackage{geometry}
\usepackage[capitalize]{cleveref}
\crefname{section}{Sec.}{Secs.}
\Crefname{section}{Section}{Sections}
\crefname{table}{Tab.}{Tabs.}
\Crefname{table}{Table}{Tables}

\usepackage[backend=biber, style=numeric, sorting=nty]{biblatex}
\title{Neural Networks at a Fraction with Pruned Quaternions}
\author{Sahel Mohammad Iqbal, Subhankar Mishra \\
National Institute of Science Education and Research, India \\
\{sahelm.iqbal, smishra\}@niser.ac.in}
\date{}

\begin{document}
\maketitle

\begin{abstract}
  Contemporary state-of-the-art neural networks have increasingly large numbers of parameters, which prevents their deployment on devices with limited computational power. Pruning is one technique to remove unnecessary weights and reduce resource requirements for training and inference. In addition, for ML tasks where the input data is multi-dimensional, using higher-dimensional data embeddings such as complex numbers or quaternions has been shown to reduce the parameter count while maintaining accuracy. In this work, we conduct pruning on real and quaternion-valued implementations of different architectures on classification tasks. We find that for some architectures, at very high sparsity levels, quaternion models provide higher accuracies than their real counterparts. For example, at the task of image classification on CIFAR-10 using Conv-4, at $3\%$ of the number of parameters as the original model, the pruned quaternion version outperforms the pruned real by more than $10\%$. Experiments on various network architectures and datasets show that for deployment in extremely resource-constrained environments, a sparse quaternion network might be a better candidate than a real sparse model of similar architecture.
\end{abstract}

\section{Introduction}
\label{sec:introduction}

A key attribute of any neural network architecture is the number of trainable parameters that it has. In general, the greater the number of model parameters, the greater its demands on computational power, time, and energy to train and perform inference. Contemporary state-of-the-art deep neural networks have model parameters that often run into tens or even hundreds of millions \cite{dsfd, Simonyan2014, bert}, imposing great demands on the hardware needed to train these models.

There are several real-world scenarios where we would like to deploy well-performing models to edge devices such as mobile phones. An example would be when dealing with private user data such as images, where sending data to a back-end datacenter for inference would be less than ideal \cite{Zhu2017}. The best (and biggest) models cannot be run on these resource-constrained computing environments \cite{mobilenet}. This leaves us with two options - either use smaller, more specialized architectures (as in MobileNet \cite{mobilenet}) or find ways to compress the bigger ones.

There are multiple compression methods by which we can reduce the resource consumption of a model such as parameter pruning and sharing, low-rank factorization and knowledge distillation \cite{Cheng2017}, but in this work, we focus on pruning.  Pruning is a method to reduce the number of parameters in a model by removing redundant weights or neurons \cite{Lecun1990}. Various studies have shown that pruning can drastically reduce model parameter counts while still maintaining accuracy \cite{Han2015, Li2017, Blalock2020}. These pruned models can also be re-trained \cite{Frankle2019}, helping to reduce resource utilization during the training stage as well.

What happens when we prune a model to extreme levels of sparsity, say 90\% or more? At this level, we typically see that the accuracy drops off \cite{Frankle2019, Han2015}, and that the pruned model can no longer match the original model. For this reason, most studies on pruning only prioritize up until the point where the pruned model is no longer on par with the parent \cite{Frankle2019, Han2015, Lecun1990, Li2017}. However, we feel that this regime is still attractive because various empirical studies have found that a large-pruned model consistently does better than a small-dense model of equal size \cite{Zhu2017, Lee2019, Gray2017}. Thus a state-of-the-art model pruned to just 2\% of its original size might still provide better accuracy than a miniature model of comparable size, even though the pruned model displays lower accuracy than the original.

Recently, another method of reducing model parameters is undergoing a surge in popularity. Using higher-dimensional data embeddings, such as complex numbers or quaternions, has been successfully shown to reduce model parameters while maintaining accuracy \cite{habertor, Trabelsi2018, Gaudet2018, Parcollet2020}. Quaternions are a 4-dimensional extension to the complex numbers introduced by the mathematician William Rowan Hamilton in 1843 \cite{Parcollet2020}, and quaternion neural networks have been built for a variety of ML tasks \cite{Zhu2018, Parcollet2018b, Gaudet2018, Parcollet2018, Pavllo2020, Comminiello2019}. Converting a real model to quaternion can lead to a 75\% reduction in model parameters (which is explained in more detail in \cref{subsec:weight_reduction}), making it a suitable method for model compression.

In the present work, we employ pruning on quaternion networks to see if they have any advantages over their real counterparts at high levels of model sparsities. To the extent of our knowledge, there are no prior studies that explore weight-reduction in neural networks by combining pruning with quaternion-valued neural networks (or any other higher-dimensional data structure). We choose multiple neural networks for image classification on the MNIST \cite{mnist}, CIFAR-10 and CIFAR-100 \cite{cifar} datasets, build equivalent quaternion representations, and conduct pruning experiments on both real and quaternion implementations. We find that at extreme sparsities (approximately 10\% or fewer parameters as the real, unpruned model), the quaternion model outperforms the real. Thus for deploying in a resource-constrained device, a quaternion pruned model might provide the best accuracy out of all available options.

Specificially, the contributions of this work can be summarized as follows. We conduct pruning experiments on real and quaternion-valued implementations of different network architectures. Through this we show that 1) the lottery ticket hypothesis \cite{Frankle2019} is valid for quaternion models, meaning that pruned quaternion models can be re-trained from scratch to the same accuracy as the unpruned model, and 2) at very high model sparsities, the quaternion equivalent displays higher accuracy than the real network.

\section{Related Work}
\label{sec:related_work}

\subsection{Pruning}
From the early 1990s, we have known that the majority of weights in a trained neural network can be pruned without sacrificing its accuracy \cite{Lecun1990, reed1993, hassibi1992}. Earlier works were done on simple architectures, but recently pruning has also been shown to work on much more extensive architectures such as VGG \cite{vgg} and ResNet \cite{resnet, Frankle2019, Blalock2020}. These studies showed that modern state-of-the-art architectures are often heavily over-parameterized and that they only require a far fewer number of parameters to learn the necessary function representations \cite{Frankle2019}.

In most studies, authors generally attempt to prune after the training process, at the end of which the weights would be ordered based on their contribution to the output. One common example of a heuristic for such ordering is the weight magnitude, where the contribution of individual weights to the output are judged based on their absolute values \cite{Han2015}. However, when training pruned networks from scratch, often they could not match the original accuracy, and the pruned models did much worse  \cite{Han2015, Li2017}. The 'Lottery Ticket Hypothesis' paper \cite{Frankle2019} showed that these pruned networks could indeed be trained from the beginning, but we had to be pair the model with the initial weights for the unpruned network. This work showed that the benefits from pruning could be realized during the training process, which could potentially reduce the resource requirements of training by a huge amount.

\begin{figure*}[htbp]
	\centering
	\includegraphics[width=0.9\textwidth]{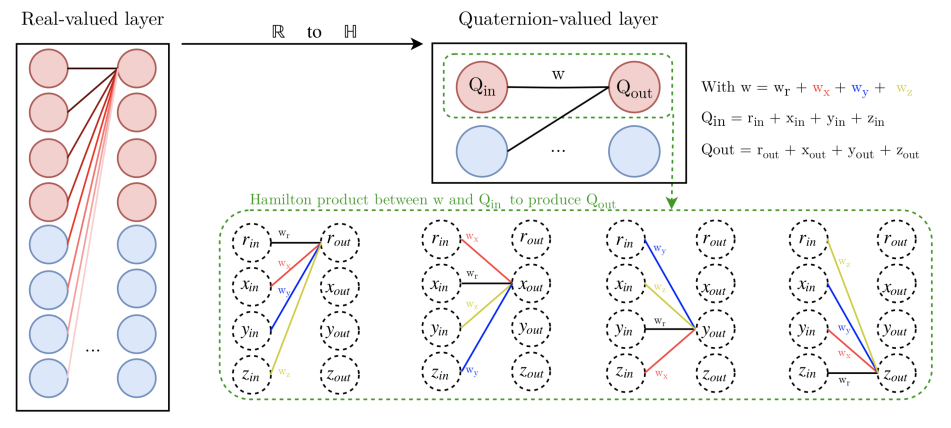}
	\caption{Depiction of how using different linear combinations of coefficients in the Hamilton product results in a reduction in the number of weights in a neural network. Image copyright Tituan Parcollet \cite{Parcollet2018b}, reproduced with permission.}
	\label{fig:quaternion_weight_sharing}
\end{figure*}

Pruning is generally of two types. In structured pruning, weights are pruned in groups by removing whole neurons or entire channels \cite{Renda2020, Li2017}. Structured pruning leads to a reduction in the size of the model and improves model inference speeds as the dimensions of weight matrices are reduced \cite{Liu2017}. Unstructured pruning, by contrast, is where individual weights of neurons are removed instead of entire neurons or channels \cite{Lecun1990, Han2015}. While unstructured pruning reduces the number of parameters, this does not immediately manifest itself through an improvement in inference speeds \cite{Park2017}. This is because unlike structured pruning, here weight matrices remain the same dimensions but are instead simply made sparse, which current hardware technology is not capable of optimizing \cite{Park2017}. However, empirical studies have shown that unstructured pruning often yields much better results than structured \cite{Renda2020}. In addition, with both pruning methods, the resource demands for training are reduced because we now have a far smaller set of weights that we need to optimize. Research to optimize sparse operations on current hardware has shown promising results \cite{Elsen2020}, and hardware that can accelerate sparse-matrix multiplications are being built \cite{nvidia}. This suggests that all the predicted theoretical performance gains from pruning could soon be realized in the near future.

An appealing property of pruned networks, and the one that justifies this work, is that in general, a large-sparse (pruned) model performs better than a small-dense (unpruned) one with an equal number of weights \cite{Blalock2020}. Multiple studies have shown that for a variety of model architectures on different tasks, sparse models consistently outperform their dense counterparts \cite{Zhu2017, Lee2019, Kalchbrenner2018, Gray2017}. This implies that given a resource constraint on the size of the models that a user can run, their best bet at achieving the greatest possible accuracy would be to use a large-pruned model rather than a small-dense one. This is applicable even when the pruned model cannot match the accuracy of the original, unpruned model.

\subsection{Quaternions}

In recent years, there has been a marked increase in works that address the question of whether it is more optimal to use multi-dimensional data embeddings in applications where the input data is multi-dimensional (refer \cite{Parcollet2020} for a comprehensive review). For example, Trabelsi et al. \cite{Trabelsi2018} demonstrated that at the task of music transcription, where the input signal is two-dimensional (consisting of magnitude and phase of the signal), complex-valued neural networks outperformed comparable real models. Complex numbers, however, are insufficient to represent higher-dimensional inputs such as the three channels of a color image, which is why some studies extended this idea to four-dimensional quaternions. Zhu et al. \cite{Zhu2018} compared quaternion and real-valued convolutional neural networks (henceforth referred to as \quat\ and \real\, respectively) with similar architectures and the same number of parameters on the CIFAR-10 dataset. They found that \quat\ achieved faster convergence on the training loss as well as higher classification accuracy on the test set compared to \real. Gaudet and Maida \cite{Gaudet2018} made a similar comparison with image classification on the CIFAR-10 and CIFAR-100 datasets and image segmentation on the KITTI Road Segmentation dataset \cite{Fritsch2013}, but this time with \quat\ having a quarter of the number of parameters as \real. They reported that on both tasks, quaternion models gave higher accuracy than real and complex networks while having a lower parameter count. Similar advantages for quaternion neural networks over real networks were also found by Parcollet et al. \cite{Parcollet2018b} for speech recognition.

\section{Theory of Quaternions}
\label{sec:quaternions}

\subsection{Quaternion Algebra}
\label{subsec:quaternion_algebra}

Quaternions are a four-dimnensional extension to the complex numbers, and a general quaternion $q$ may be written as
\begin{equation}
	q = r + x\textbf{i} + y\textbf{j} + z\textbf{k}
	\label{eq:quaternion}
\end{equation}
where $r, x, y, z \in\mathbb{R}$ and $\textbf{i}$, $\textbf{j}$ and $\textbf{k}$ are complex entities which follow the relations
\begin{equation}\label{eq:quaternion_relation}
	\textbf{i}^2 = \textbf{j}^2 = \textbf{k}^2 = \textbf{i}\textbf{j}\textbf{k} = -1
\end{equation}
Given two quaternions $q_1$ and $q_2$, their product (known as the Hamilton product) is given by
\begin{align}
	q_1 \otimes q_2 = \textrm{ } & (r_1r_2 - x_1x_2 - y_1y_2 - z_1z_2) \nonumber \\
				     & (r_1x_2 + x_1r_2 + y_1z_2 - z_1y_2) \textrm{ }\textbf{i} \nonumber \\
				     & (r_1y_2 - x_1z_2 + y_1r_2 - z_1x_2) \textrm{ }\textbf{j} \nonumber \\
				     & (r_1z_2 + x_1y_2 - y_1x_2 - z_1r_2) \textrm{ }\textbf{k}
	\label{eq:hamilton_product}
\end{align}
Unlike real and complex multiplication, quaternion multiplication is not commutative. Quaternions can be represented as $4*4$ real matrices such that the matrix multiplication between such representations are consistent with the Hamilton product.
\begin{equation}
	q = \begin{bmatrix}
		r & -x & -y & -z \\
		x & r & -z & y \\
		y & z & r & -x \\
		z & -y & x & r
	\end{bmatrix}
	\label{eq:quaternion_matrix_form}
\end{equation}
This is the representation that is used to calculate quaternion products in a quaternion-valued neural network.

\begin{table*}[htbp]
  \centering
  \footnotesize
  \begin{tabular}{@{}lcccc@{}}
    \toprule
    Model & Lenet-300-100 & Conv-2 & Conv-4 & Conv-6\\
    \midrule
    & & & CIFAR-10 & CIFAR-10 \\
    Datasets & MNIST & CIFAR-10 & CIFAR-100 & CIFAR-100 \\ \midrule
            & & & & $64*64$, pool \\
            & & & $64*64$, pool & $128*128$, pool \\
    Conv Layers & & $2*64$, pool & $128*128$, pool & $256*256$, pool \\ \midrule
    FC Layers & 300, 100, 10 & 256, 256, 10/100 & 256, 256, 10/100 & 256, 256, 10/100 \\ \midrule
    All/Conv Weights (Real) & 266.6K & 4.30M/38K & 2.42M/260K & 2.26M/1.14M \\ \midrule
    All/Conv Weights (Quat) & 67.7K & 1.08M/9.9K & 609K/65K & 569K/287K \\ \midrule
    Training epochs/Batch size & 40/60 & 40/60 & 40/60 & 60/60 \\ \midrule
    Optimizer/Learning rate & Adam/1.2e-3 & Adam/2e-4 & Adam/3e-4 & Adam/3e-4 \\
    \bottomrule
  \end{tabular}
  \caption{Model architectures, datasets and hyper-parameters tested in this paper. The number of weights for Conv-2,4,6 are reported for CIFAR-10 classification. Architectures and hyper-parameters have been kept the same as those used in \cite{Frankle2019} to facilitate a direct comparison.}
  \label{tab:model_details}
\end{table*}

\subsection{How does weight-reduction happen?}
\label{subsec:weight_reduction}

Consider a small section of a fully-connected neural network with four input and four output neurons. In the case of a real-valued implementation, this layer would require $4*4=16$ weights. However, if we were to view the above network as consisting of one input quaternion and one output quaternion, then using the Hamilton product, we would only require one quaternion weight, or four real weights, to connect them. Thus provided the number of neurons in all layers are divisible by 4, we can obtain a $75\%$ reduction in the number of parameters of a network by converting it to a quaternion implementation. This is shown graphically in \cref{fig:quaternion_weight_sharing}. This weight-reduction is explained in greater detail in \cite{Parcollet2018b, Parcollet2020}.

\section{Methodology}
\label{sec:methodology}

Our experiments were run on classification tasks on the MNIST \cite{mnist}, CIFAR-10 and CIFAR-100 \cite{cifar} datasets. We chose to demonstrate our experiments at image classification tasks following the lead of some of the most important works in network compression \cite{Frankle2019, frankle2021} so that our results may be contrasted with state-of-the-art. We used the fully-connected Lenet-300-100 architecture from \cite{Lecun1998}, and the Conv-2, Conv-4, and Conv-6 convolutional models from \cite{Frankle2019}. Complete details about model architectures and hyper-parameters are given in \cref{tab:model_details}.

Our goal in this work was to compare pruning on real and quaternion-valued implementations of different architectures. To do this, we first constructed quaternion equivalents of every model with the condition that they both have the same number of real neurons. \quat\ would thus have one-fourth the number of quaternion neurons (since four real neurons constitute a quaternion neuron) and one-fourth the number of weights as \real\ (because of the weight-sharing property of the Hamilton product explained in \cref{subsec:weight_reduction}). We used a real output layer for \quat\ because for the MNIST and CIFAR-10 datasets, the number of output labels (10) is not divisible by 4. An alternative option here was to use 10 quaternion neurons for the output layer and then take their norms. We chose not to do this because this would break the equality condition that we just stated, that the number of real neurons across implementations be equal. All other network layers were replaced by equivalent quaternion implementations.  The hyper-parameters used for training are also the same as \cite{Frankle2019} in order to make direct comparisons.

For the MNIST dataset, since the images are grayscale and have only one channel, the input image is flattened and each set of four pixels are fed to a quaternion neuron of Lenet-300-100. On the other hand, for color vision tasks such as classification on the CIFAR-10, it makes more sense to treat the RGB channels of each pixel as belonging to a single quaternion neuron. To do this, we need to add one more channel to the input images, and the equality condition has to be relaxed for the input channel. A few different options exist, such as a channel with all 0s or using an additional layer to learn the fourth channel \cite{Gaudet2018}. For our implementation, we chose to use the grayscale values of the input image as the additional channel.

For pruning experiments, we use iterative pruning, where the model is pruned and then re-trained for a certain number of epochs over multiple iterations. We prune the networks using a global pruning technique with $20\%$ of the weights in the model pruned in each pruning iteration. We chose global pruning over layer-wise pruning because global-pruning can find smaller lottery tickets for larger networks \cite{Frankle2019}, and we wanted to keep the methodology consistent across all the architectures that we test. We only prune weights and exempt biases, as biases constitute only a tiny proportion of the total parameters in a model. At each level of model sparsity, the pruned model is re-trained from scratch using the initial weights for the same duration as the original model (as done in \cite{Frankle2019}), and it is the accuracy of these re-trained pruned models that we have reported throughout this paper. Our emphasis is on re-trainable pruned models because our primary concern is with the practical benefits of pruning conferred during training. The models are pruned until their accuracies drop below a threshod (which we chose to be $30\%$) for two successive pruning iterations. This was done to save computational resources by preventing pruning the model beyond the point where it is of any practical use.

Pruning experiments were conducted using PyTorch \cite{pytorch}. PyTorch implementations of the various operations and building blocks necessary to build quaternion convolutional neural networks have been borrowed from the hTorch library (MIT License) \cite{htorch}. This library uses a split activation function \cite{Arena1994} where ReLU \cite{Glorot2010} is applied individually to the four different components of each quaternion. The back-propagation algorithm employed for quaternions is a generalization of those for real and complex networks \cite{Nitta1995, Parcollet2018, Trabelsi2018}. All experiments were carried out on a single Nvidia RTX 3090 GPU.

\section{Experimental Results}
\label{sec:experimental_results}

\begin{figure*}[htbp]
  \centering
  \begin{subfigure}{\figsize}
	  \centering
    \includegraphics[width=\subfigsize]{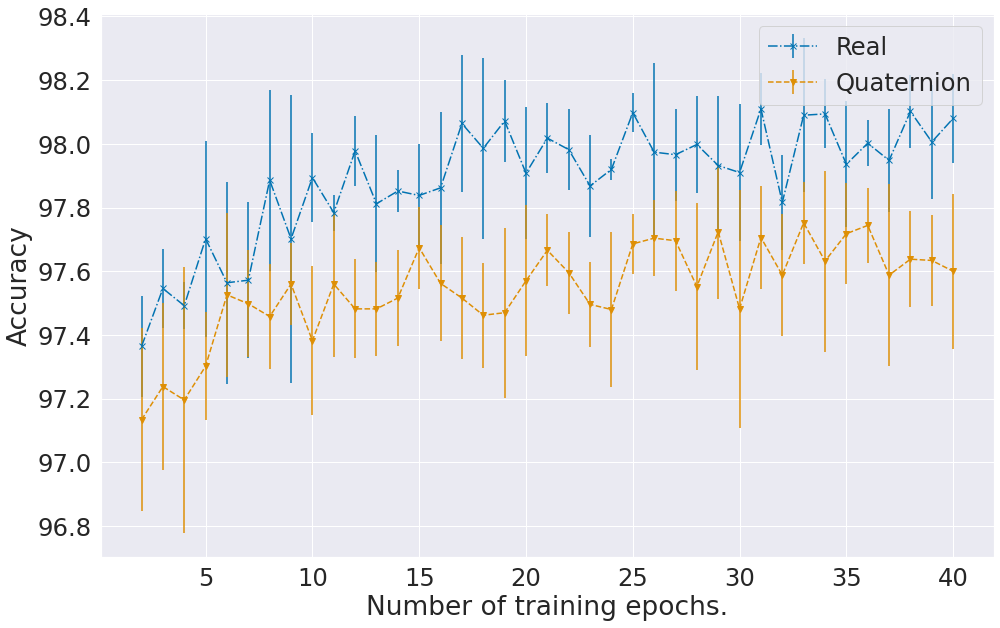}
    \caption{Lenet-300-100 on MNIST}
    \label{fig:lenet_300_100_1}
  \end{subfigure}
  \begin{subfigure}{\figsize}
	  \centering
    \includegraphics[width=\subfigsize]{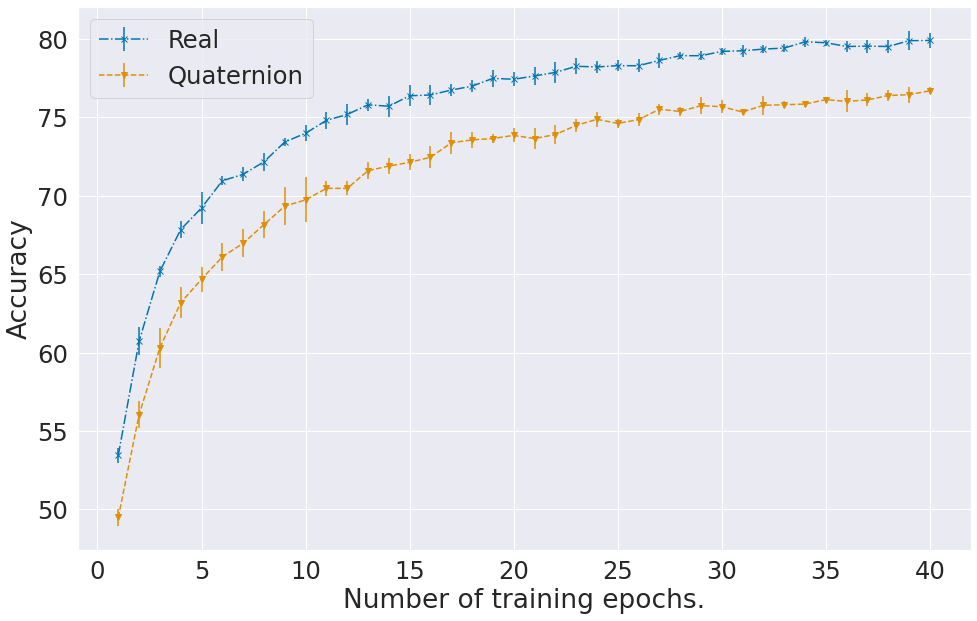}
    \caption{Conv-2 on CIFAR-10}
    \label{fig:conv_2_1}
  \end{subfigure}
  \begin{subfigure}{\figsize}
	  \centering
    \includegraphics[width=\subfigsize]{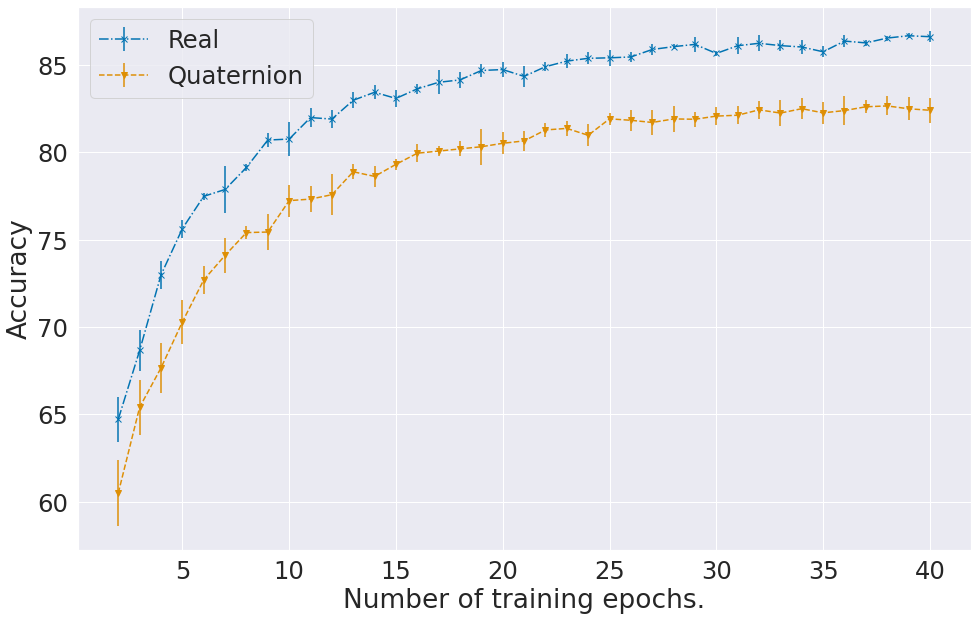}
    \caption{Conv-4 on CIFAR-10}
    \label{fig:conv_4_1}
  \end{subfigure}
  \begin{subfigure}{\figsize}
	  \centering
    \includegraphics[width=\subfigsize]{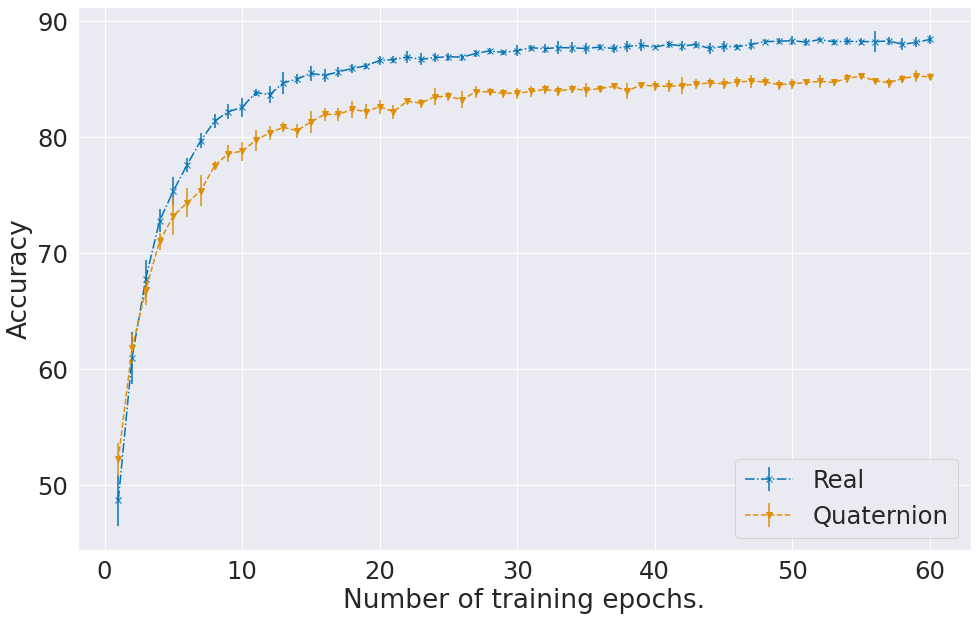}
    \caption{Conv-6 on CIFAR-10}
    \label{fig:conv_6_1}
  \end{subfigure}
  \begin{subfigure}{\figsize}
	  \centering
    \includegraphics[width=\subfigsize]{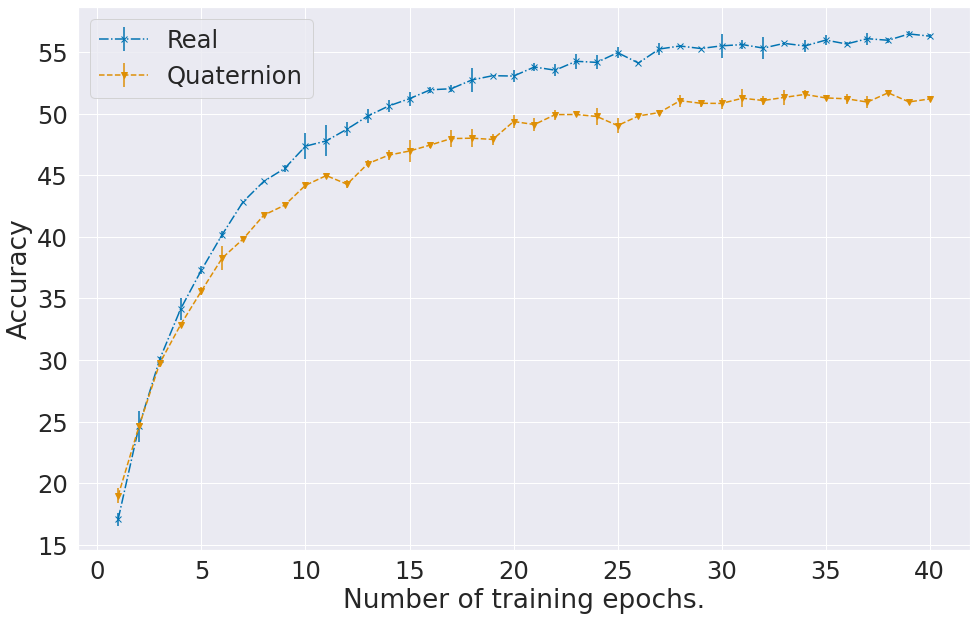}
    \caption{Conv-4 on CIFAR-100}
    \label{fig:conv_4_cifar100_1}
  \end{subfigure}
  \begin{subfigure}{\figsize}
	  \centering
    \includegraphics[width=\subfigsize]{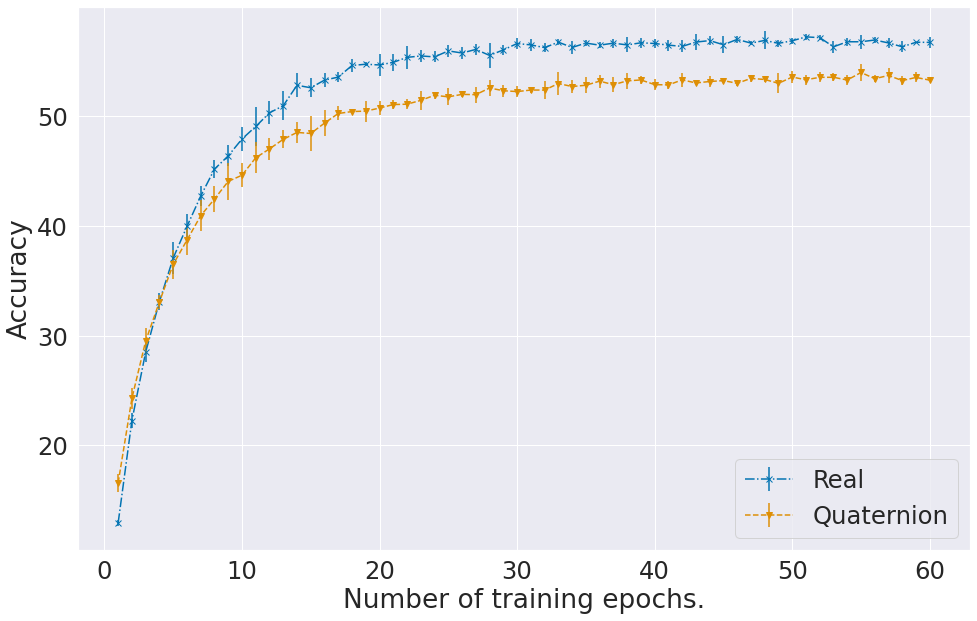}
    \caption{Conv-6 on CIFAR-100}
    \label{fig:conv_6_cifar100_1}
  \end{subfigure}
  \caption{Training results for various architectures for real and quaternion implementations. Results are the mean over 5 trials, and the error bars are the standard deviation.}
  \label{fig:training_1}
\end{figure*}

A few previous studies using quaternions had found that at certain tasks, quaternions can outperform real networks while having an equal or smaller number of parameters, including for computer vision tasks such as classification which we consider in this paper \cite{Zhu2018, Gaudet2018}. In our experiments, where \quat\ has a quarter of the number of parameters as \real, we found that \quat\ cannot, in general, be trained to the same accuracy as \real\ within a fixed number of iterations. The training results for real and quaternion implementations of the various models are reported in \cref{fig:training_1}. The relative accuracy difference between real and quaternion implementations are different for different models. We also found that this varied depending on the hyper-parameters used, and so it may be possible that for some set of hyper-parameters, \quat\ could be made as accurate as \real, reproducing the results given in the references mentioned earlier. We were unable to find this subset in our experimentation.

\begin{figure*}
  \centering
  \begin{subfigure}{\figsize}
	  \centering
    \includegraphics[width=\subfigsize]{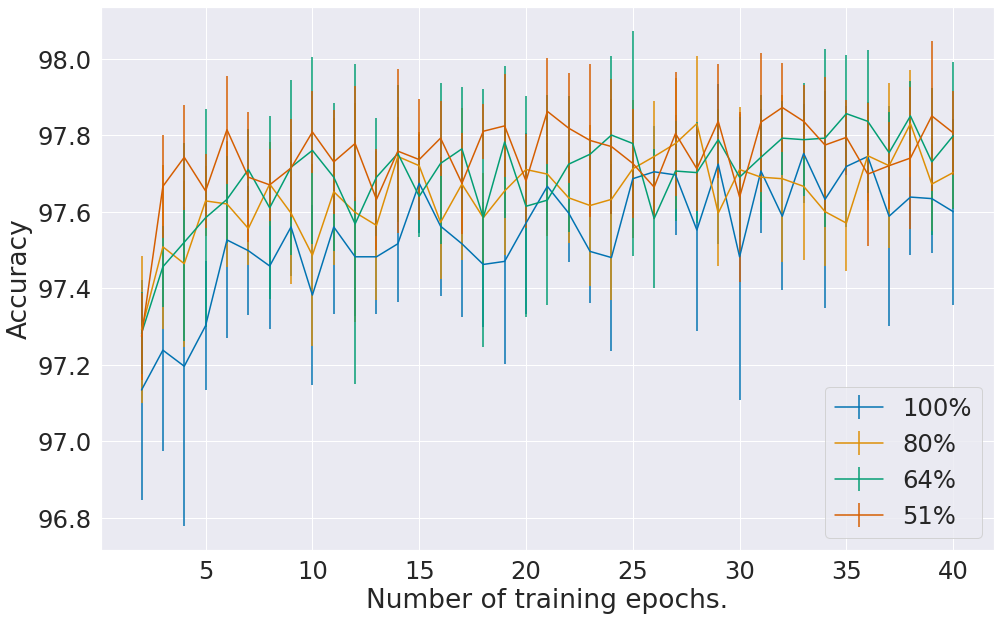}
    \caption{Lenet-300-100 on MNIST}
    \label{fig:lenet_300_100_0}
  \end{subfigure}
  \begin{subfigure}{\figsize}
	  \centering
    \includegraphics[width=\subfigsize]{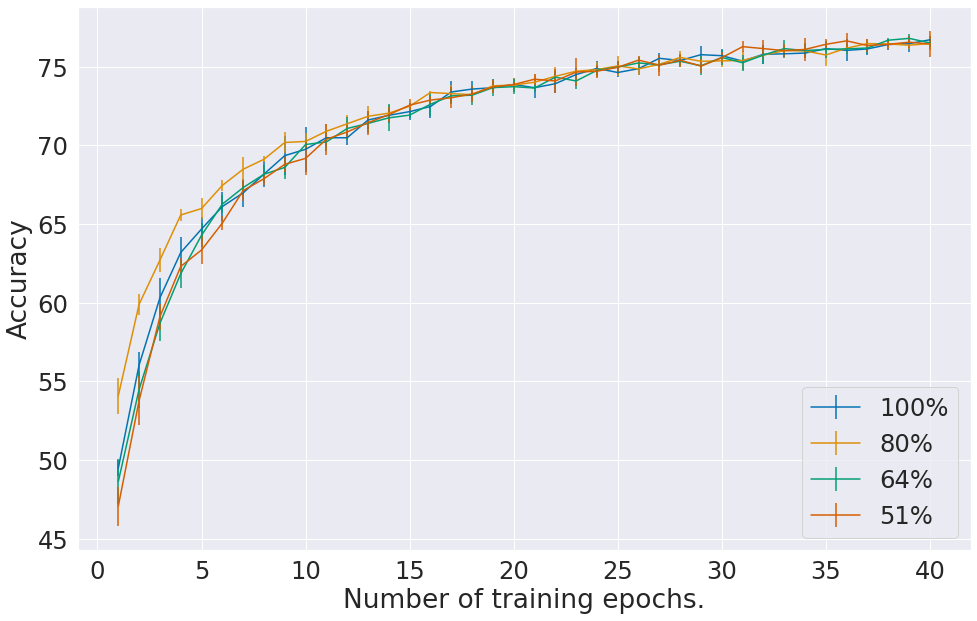}
    \caption{Conv-2 on CIFAR-10}
    \label{fig:conv_2_0}
  \end{subfigure}
  \begin{subfigure}{\figsize}
	  \centering
    \includegraphics[width=\subfigsize]{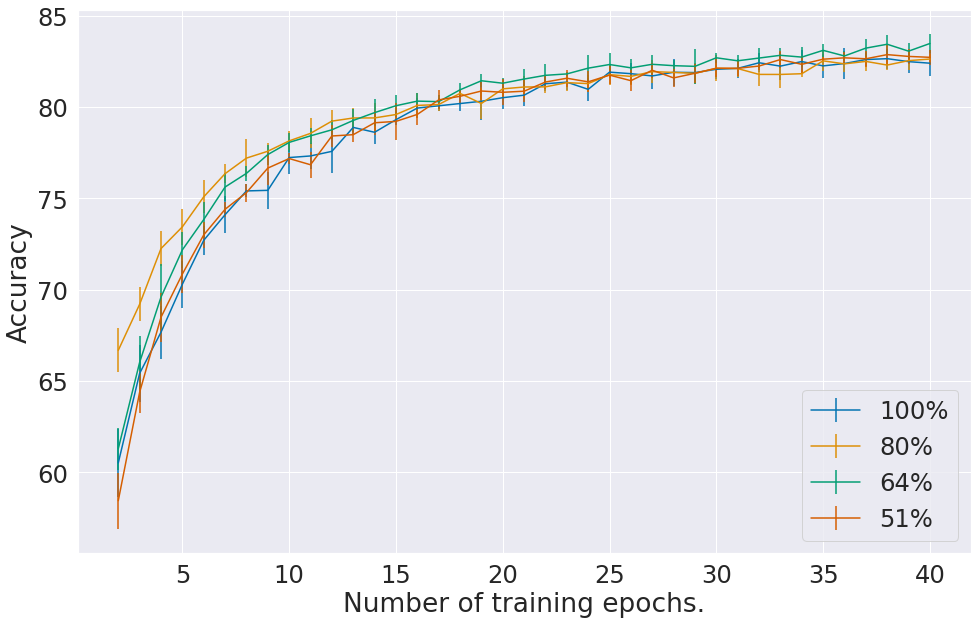}
    \caption{Conv-4 on CIFAR-10}
    \label{fig:conv_4_0}
  \end{subfigure}
  \begin{subfigure}{\figsize}
	  \centering
    \includegraphics[width=\subfigsize]{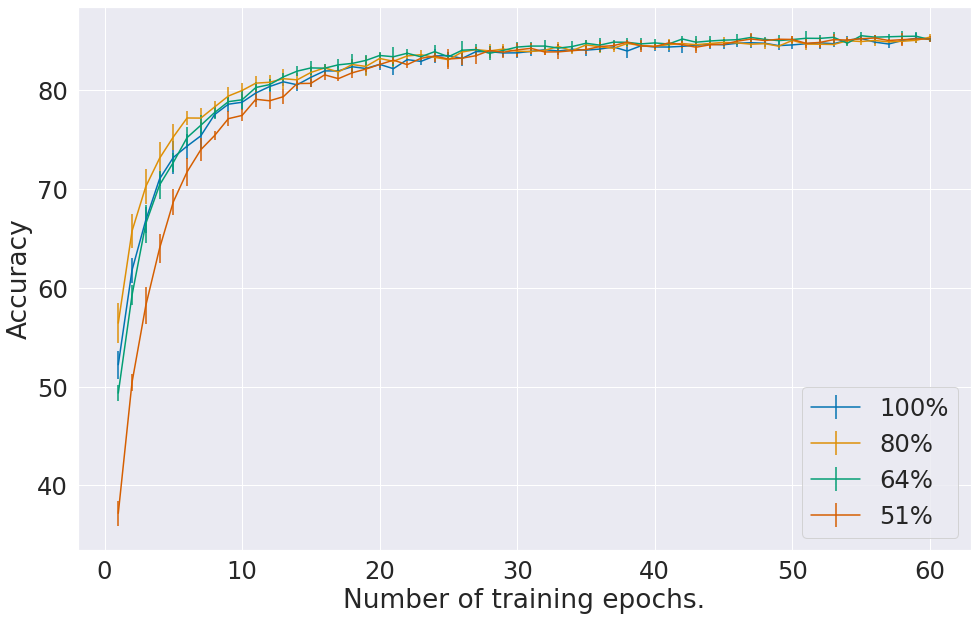}
    \caption{Conv-6 on CIFAR-10}
    \label{fig:conv_6_0}
  \end{subfigure}
  \begin{subfigure}{\figsize}
	  \centering
    \includegraphics[width=\subfigsize]{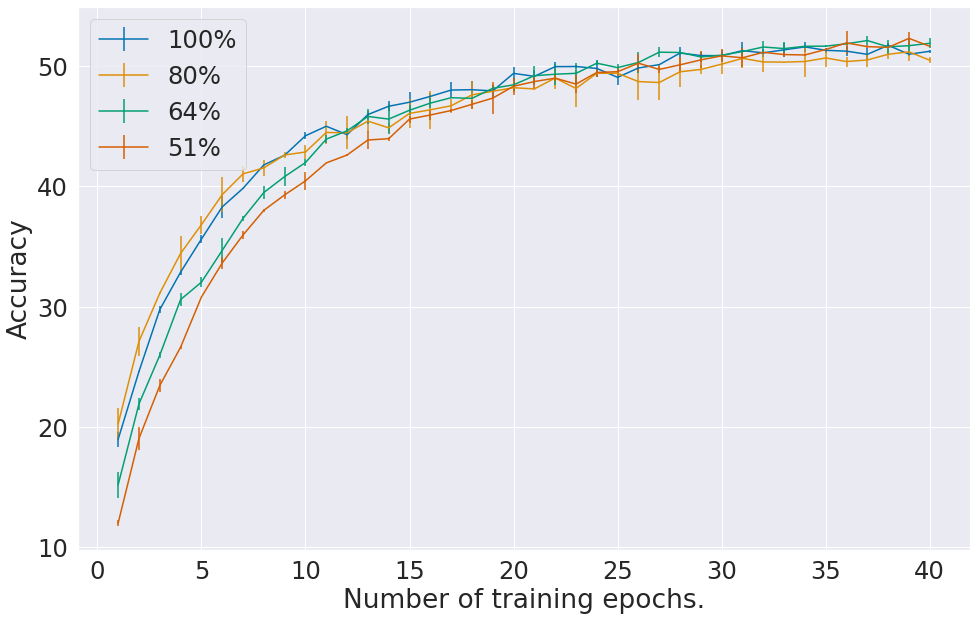}
    \caption{Conv-4 on CIFAR-100}
    \label{fig:conv_4_cifar100_0}
  \end{subfigure}
  \begin{subfigure}{\figsize}
	  \centering
    \includegraphics[width=\subfigsize]{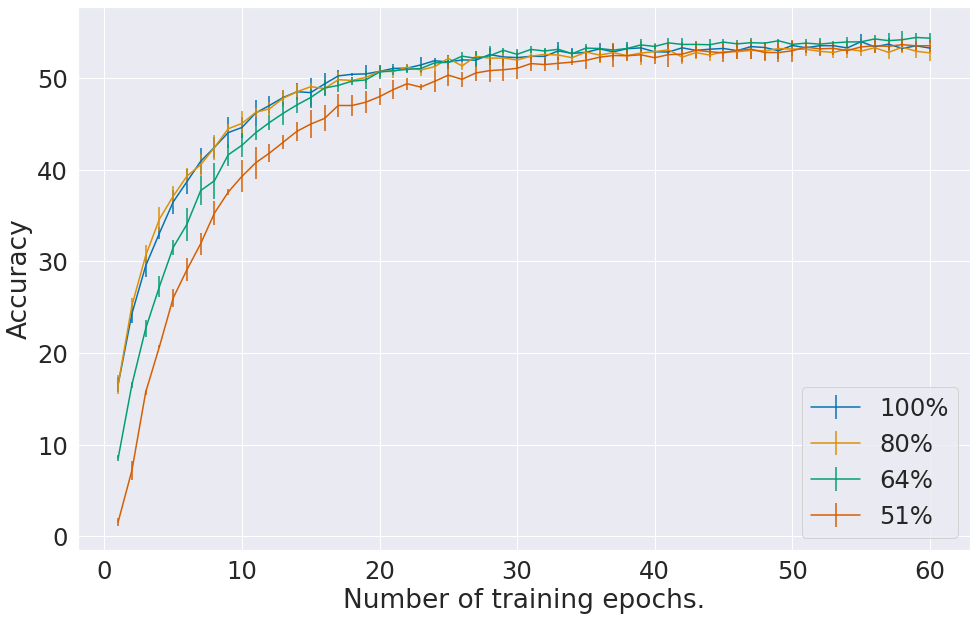}
    \caption{Conv-6 on CIFAR-100}
    \label{fig:conv_6_cifar100_0}
  \end{subfigure}
  \caption{Training results for quaternion implementations of models at different sparsities. Pruned models have been re-trained from scratch with the intial weights. Plot labels are the percentage of weights remaining after pruning. Results are the mean over 5 trials, and the error bars are the standard deviation.}
  \label{fig:lth_quat}
\end{figure*}

\begin{figure*}
  \centering
  \begin{subfigure}{\figsize}
	  \centering
    \includegraphics[width=\subfigsize]{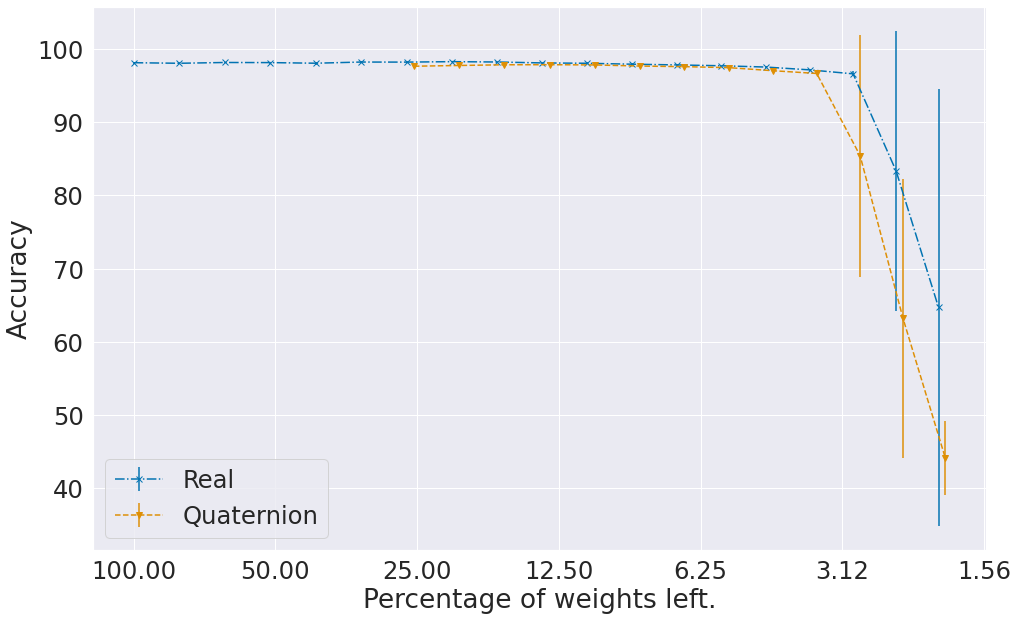}
    \caption{Lenet-300-100 on MNIST}
    \label{fig:lenet_300_100_2}
  \end{subfigure}
  \begin{subfigure}{\figsize}
	  \centering
    \includegraphics[width=\subfigsize]{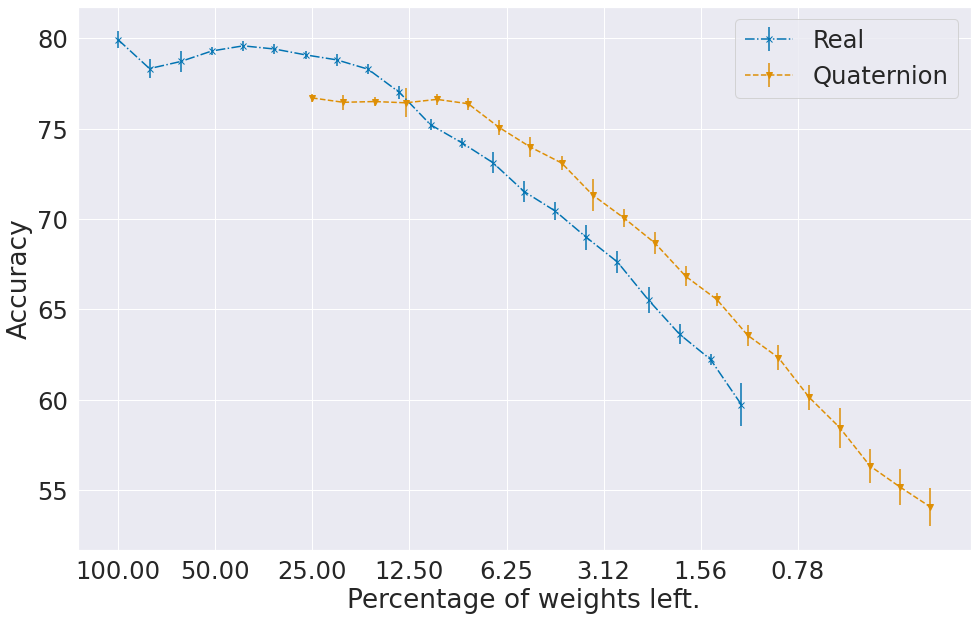}
    \caption{Conv-2 on CIFAR-10}
    \label{fig:conv_2_2}
  \end{subfigure}
  \begin{subfigure}{\figsize}
	  \centering
    \includegraphics[width=\subfigsize]{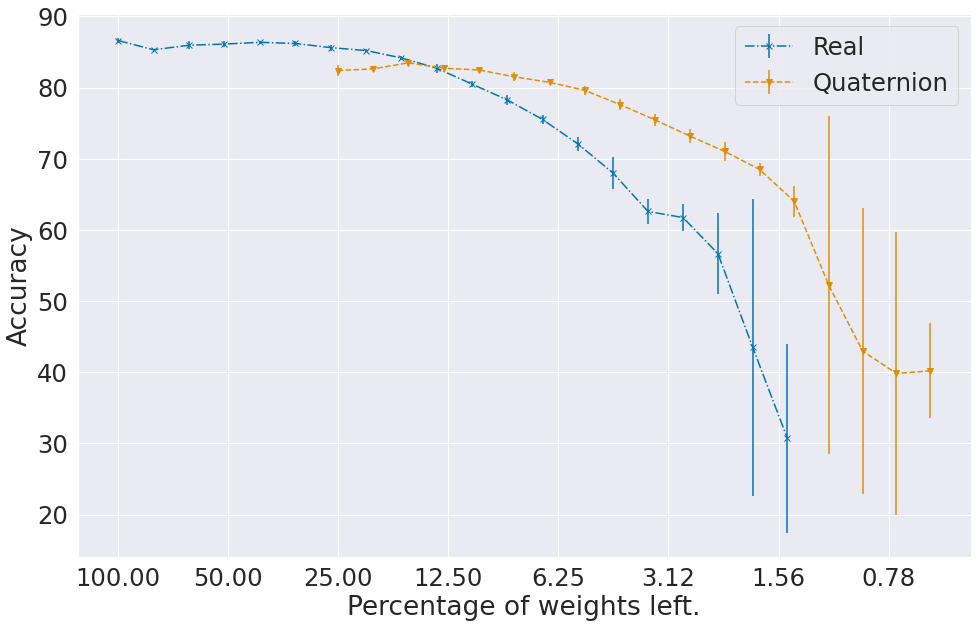}
    \caption{Conv-4 on CIFAR-10}
    \label{fig:conv_4_2}
  \end{subfigure}
  \begin{subfigure}{\figsize}
	  \centering
    \includegraphics[width=\subfigsize]{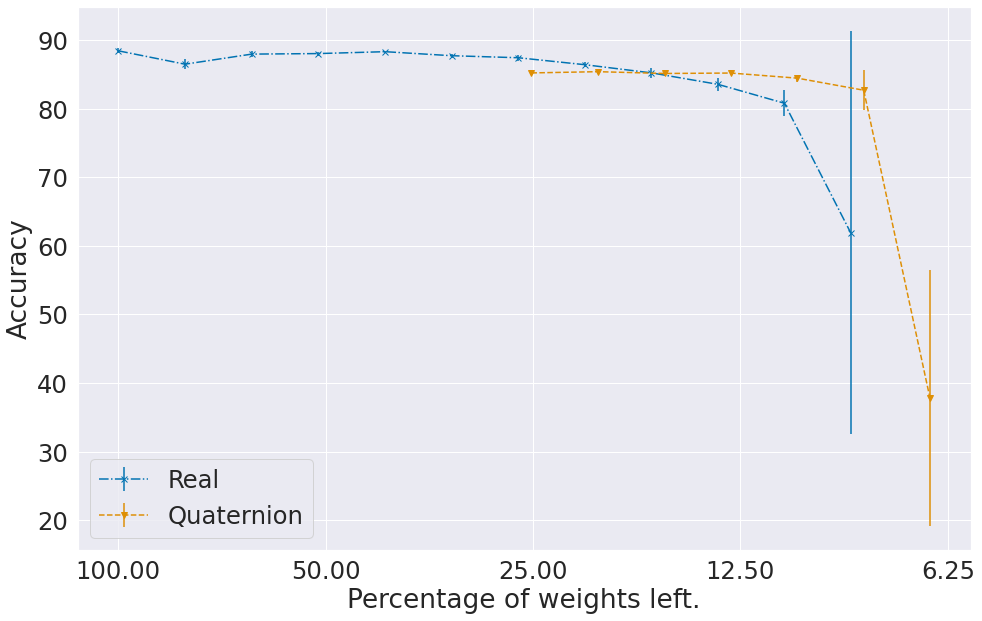}
    \caption{Conv-6 on CIFAR-10}
    \label{fig:conv_6_2}
  \end{subfigure}
  \begin{subfigure}{\figsize}
	  \centering
    \includegraphics[width=\subfigsize]{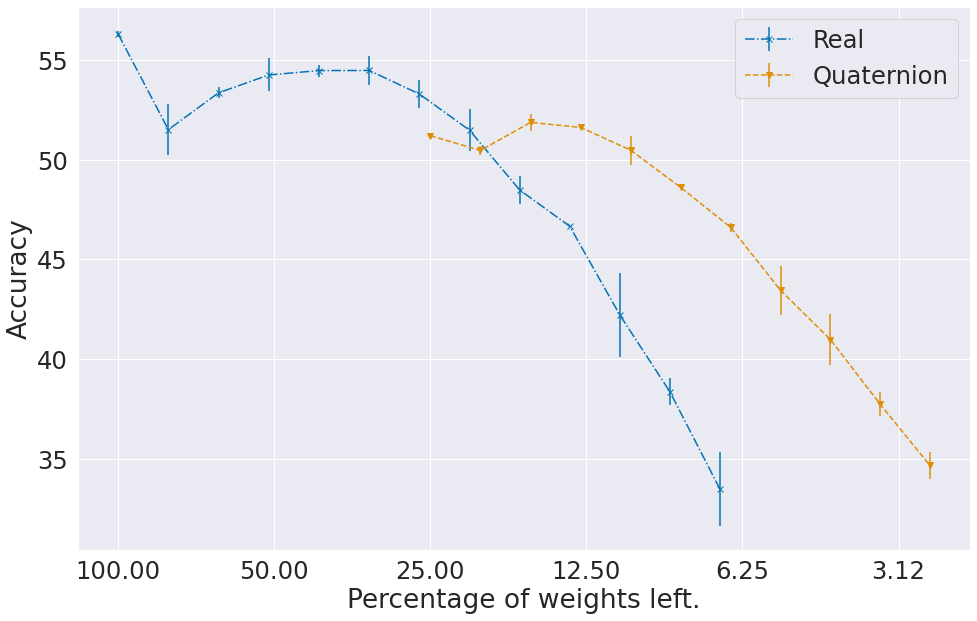}
    \caption{Conv-4 on CIFAR-100}
    \label{fig:conv_4_cifar100_2}
  \end{subfigure}
  \begin{subfigure}{\figsize}
	  \centering
    \includegraphics[width=\subfigsize]{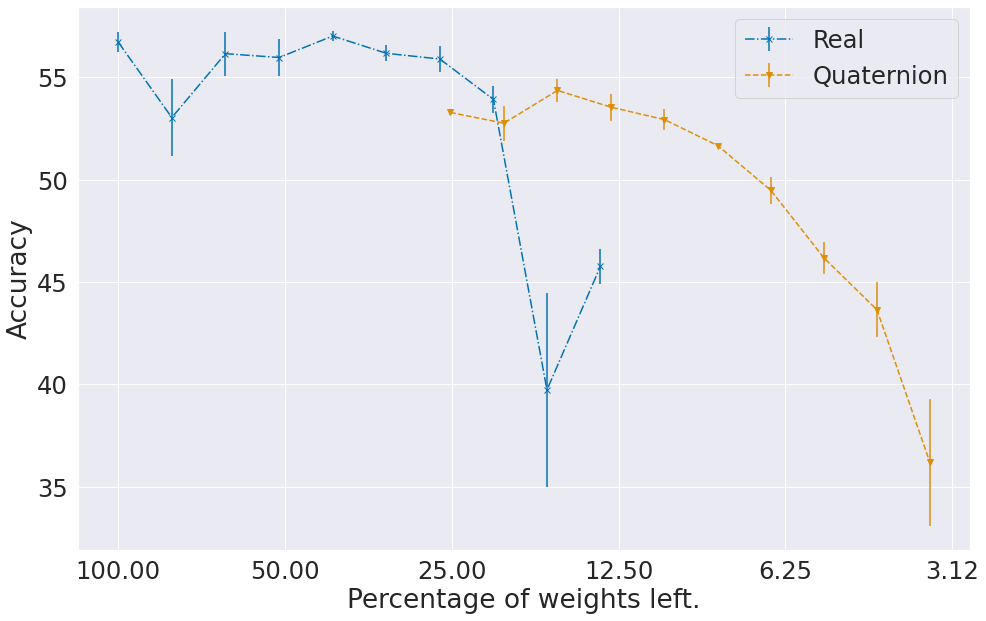}
    \caption{Conv-6 on CIFAR-100}
    \label{fig:conv_6_cifar100_2}
  \end{subfigure}
  \caption{Accuracy vs sparsity results for real and quaternion implementations of various architectures. Quaternion curves start at around the $25\%$ mark because unpruned \quat\ have only one-fourth (approx.) the number of parameters as the corresponding unpruned \real. Models are pruned until their accuracies drop below $30\%$ for two successive pruning iteraions.}
  \label{fig:pruning_1}
\end{figure*}

On re-training pruned models from scratch, we found that just like \real, pruned \quat\ are capable of matching or exceeding the accuracy of the original, unpruned model (\cref{fig:lth_quat}). For example, in \cref{fig:lenet_300_100_0}, a pruned model with $51\%$ weights ends training with higher accuracy than the unpruned model. This result held for all six architecture-dataset pairs that we tested. This shows that the lottery ticket hypothesis is valid for quaternion-valued networks as well.

The accuracies of pruned \quat\ and \real\ at various model sparsities are given in \cref{fig:pruning_1}. On examining this figure, certain patterns become evident. Consider Conv-4 on CIFAR-10 (\cref{fig:conv_4_2}) as an illustrative example. The curve for \quat\ starts at the $25\%$ mark as by construction, \quat\ only has that many parameters compared to \real. \quat\ also starts out with lower accuracy than \real, which can also be seen in \cref{fig:conv_4_1}. The regime we are interested in is that of high pruning rates, at around $10\%$ of total weights or lower. Though both models start out at different accuracies, as we get to the $12.5\%$ mark, the curves for \real\ and \quat\ coincide, meaning that at this sparsity level, both perform equally well. On further pruning, the \real\ curve drops below the \quat\ curve, implying that the quaternion outperforms the real model at very high sparsity levels. At about $3.12\%$ sparsity, \quat\ shows close to $75\%$ accuracy while it is around $62\%$ for \real, a difference of more than $10\%$. This overall pattern is also repeated for all but one of the model-dataset pairs that we tested, the sole exception being Lenet-300-100 on MNIST (\cref{fig:lenet_300_100_2}).

\begin{figure*}[htbp]
  \centering
  \begin{subfigure}{\figsize}
	  \centering
    \includegraphics[width=\subfigsize]{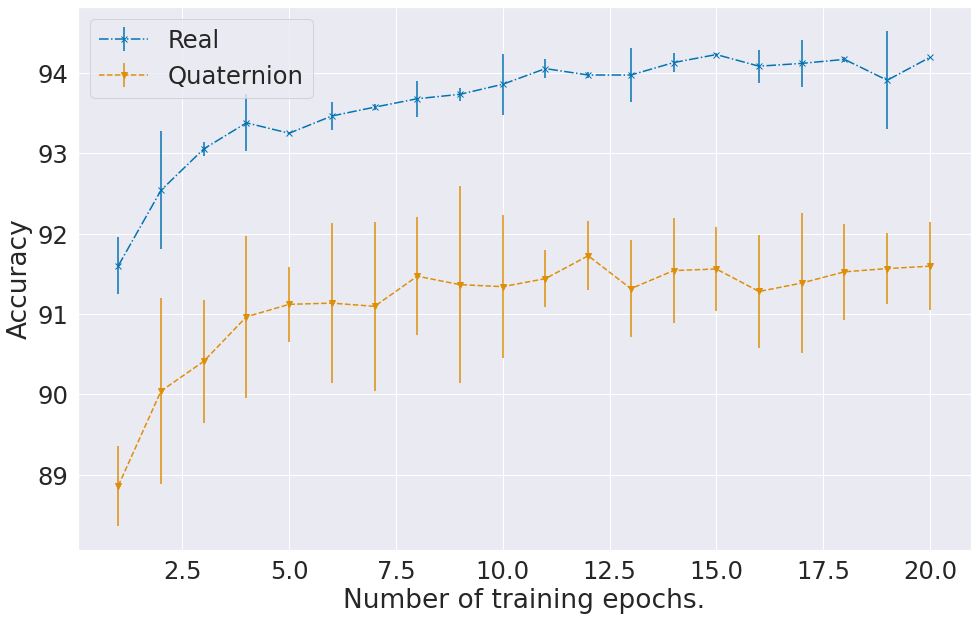}
    \caption{Training results.}
    \label{fig:lenet_12_0}
  \end{subfigure}
  \begin{subfigure}{\figsize}
	  \centering
    \includegraphics[width=\subfigsize]{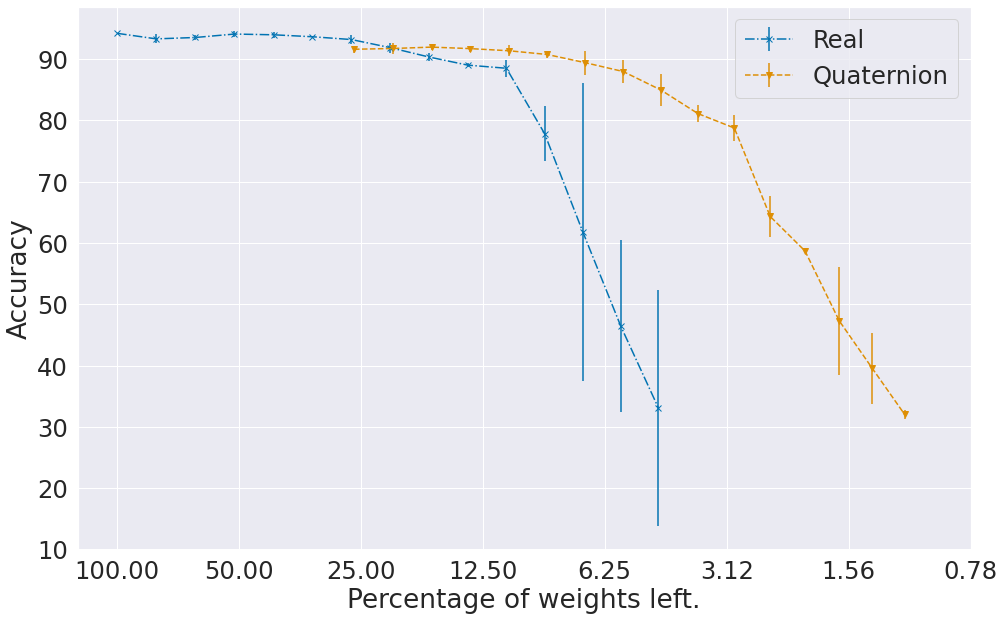}
    \caption{Pruning results.}
    \label{fig:lenet_12_2}
  \end{subfigure}
  \caption{Training and pruning results for Lenet-12 on MNIST.}
  \label{fig:lenet_12}
\end{figure*}

For Lenet-300-100, however, \real\ outperforms \quat\ at every level of sparsity. This model is different from the other three models that we tested in two ways: 1) it is a fully-connected network (no convolutional layers), and 2) it displays around $98\%$ accuracy on its task, which is considerably higher than any of the other model-dataset pairs. To isolate which of these two properties led to the difference in the sparsity-accuracy trend, we ran pruning experiments on a custom Lenet-12 model (which has a single hidden layer with 12 real neurons), which is also a fully-connected model but with lower accuracy. The results for this model on the MNIST dataset are given in \cref{fig:lenet_12}. Here the earlier trend reappears, and \quat\ performs better than \real\ at high sparsity levels. Thus Lenet-300-100 showed divergent behavior not because it is a fully-connected network but because it has very high accuracy at its task. A possible explanation for this may be that this model is so over-parameterized that the real model can be pruned to a great extent without a significant drop in accuracy. This explanation, however, cannot justify why \quat\ underperforms \real\ beyond the $4\%$ sparsity region for Lenet-300-100.

On the whole, this analysis shows that, in general, at extreme model sparsities, quaternion models perform better than their real counterparts. Although quaternion models start out at a disadvantage because of their lower initial accuracy, as the models are reduced in size, their relative performance gap diminishes until at a certain point the real model dips below the quaternion, where it remains for the rest of the pruning process.

\subsection{Using Early Stopping}
\label{subsec:early_stopping}

\begin{figure}[htbp]
  \centering
  \includegraphics[width=0.8\linewidth]{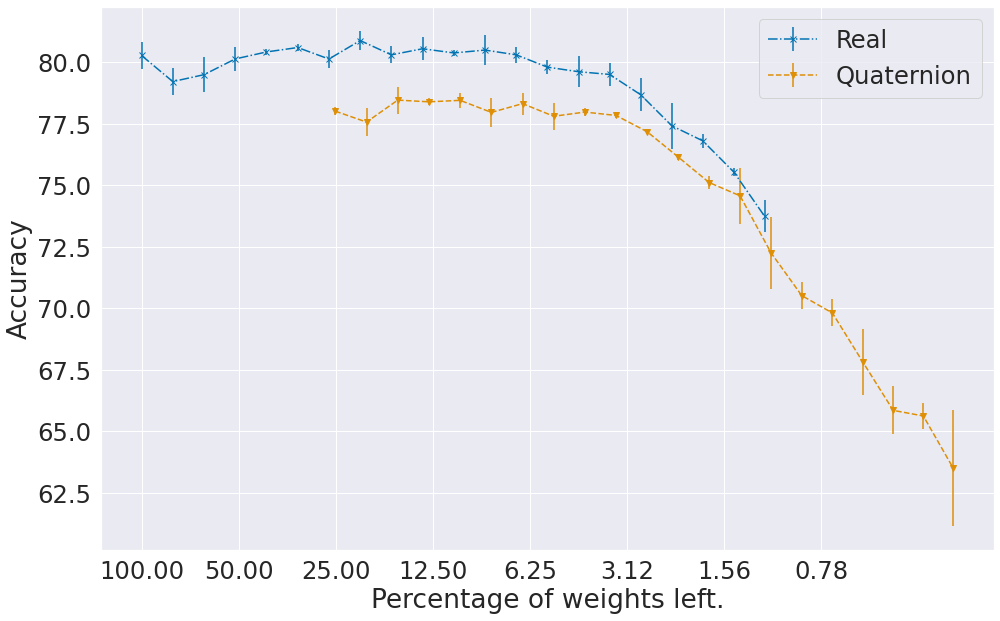}
  \caption{Accuracy vs sparsity results for Conv-2 on CIFAR-10 when using the early stopping criterion.}
  \label{fig:conv_2_early_stopping}
\end{figure}

In \cite{Frankle2019}, the authors use an early stopping criterion to stop training. The criterion used is that of the minimum validation loss. We repeated our experiments with the same early stopping criterion (with a patience of 10 iterations), but here we saw that \quat\ no longer did better than \real\ at any sparsity level for any of the models that we tested. For example, the pruning results for Conv-2 when run with the early stopping criterion is given in \cref{fig:conv_2_early_stopping}, which should be compared with \cref{fig:conv_2_2}.

\subsection{Larger Models}
\label{subsec:larger_models}

In addition to the models considered earlier, we also ran similar experiments on Resnet-18 \cite{resnet} and VGG-16 \cite{vgg}. These are deeper neural network architectures that require batch normalization layers. The first implementation of a quaternion batch normalization algorithm was given by \cite{Gaudet2018}. This implementation treats a quaternion as a single entity and requires complex matrix operations to calculate the mean and variance of each layer, and is thus extremely computationally intensive (adding a single batchnorm layer to Conv-2 increased the time taken for each trainging iteration by approximately 40 times compared to baseline). Hence we had to opt for another implementation of batch normalization given in \cite{Yin2019}. This implementation treats each individual components of quaternions separately, and is hence much faster. This obviously comes at the cost of compromising the relationships between the individual components of quaternions, but since we are already doing this with our use of the split activation function, it may not lead to any further disadvantage.

Neither Resnet nor VGG follow the trend we saw for the smaller models, in that for both of these network architectures \real\ has greater accuracy than \quat\ at all sparsity levels. Whether it is the introduction of the batchnorm layer or the depth of the architectures that is causing this reversal in trend is unclear and needs to be investigated further.

\section{Conclusions}
\label{sec:conclusions}

In this work, we conduct extensive pruning experiments on real and quaternion-valued implementations of different neural network architectures with the objective of checking whether using quaternions provides any advantages in model compression. We first found that pruned quaternion models can be re-trained from scratch to match the original accuracy of the unpruned model, showing that lottery tickets exist for quaternion networks as well. More importantly, our experiments demonstrate that when pruned to high levels of sparsities, quaternion implementations of certain models outperform their complementary real-valued models of equivalent architectures. Hence for ML tasks with multi-dimensional inputs that need to be run on devices with limited computational power, a pruned quaternion model may be a more suitable option than an analogous real network.

\subsection{Limitations and Future Work}

Like all empirical studies, the primary limitation of this work is in the scope of vision tasks, datasets and models tested. Here we tested six architectures on three different datasets at the task of classification. An extension to this work, and one that could further generalize the conclusions reached, is to consider a larger set of models and datasets, and test them on additional vision tasks such as semantic segmentation. Investigating how pruned quaternion implementations of deeper architectures that require batch normalization layers can be made to outperform their real counterparts is also identified as future work.

\newpage
\printbibliography

\end{document}